\documentclass[a4paper, 10pt, conference]{ieeeconf}
\IEEEoverridecommandlockouts

\usepackage{graphicx}
\usepackage{adjustbox}
\usepackage{multirow}
\usepackage{amssymb}
\usepackage{color}
\usepackage{tikz}
\usepackage{lineno}
\usepackage{subfig}
\usepackage{tabularx, rotating, caption}
\usepackage{array}
\usepackage{siunitx}
\usepackage{url}
\usepackage{booktabs}
\usepackage{comment}
\usepackage{mathtools}
\usepackage{pifont}%
\usepackage{cite}
\usepackage[normalem]{ulem}

\usepackage{verbatim}

\newcommand \blue[1]{\textcolor{blue}{#1}}
\newcommand \red[1]{\textcolor{red}{#1}}

\newcommand{\michele}[1]{\colorbox{purple}{\color{white}   \textsf{\textbf{michele}}} \textcolor{purple}{#1}}

\usepackage[hidelinks]{hyperref}

\begin{document}

\title{\LARGE \bf Enhancing Door--Status Detection for Autonomous Mobile Robots\\ during Environment--Specific Operational Use 
}

\author{Michele Antonazzi, Matteo Luperto, Nicola Basilico, N. Alberto Borghese%
\thanks{All authors are with the Department of Computer Science, University of Milan, Milano, Italy {\tt\small~name.surname@unimi.it}
}
}

\maketitle

\begin{abstract}

Door--status detection, namely recognising the presence of a door and its status (open or closed), can induce a remarkable impact on a mobile robot's navigation performance, especially for dynamic settings where doors can enable or disable passages, changing the topology of the map. In this work, we address the problem of building a door--status detector module for a mobile robot operating in the same environment for a long time, thus observing the same set of doors from different points of view. First, we show how to improve the mainstream approach based on object detection by considering the constrained perception setup typical of a mobile robot. Hence, we devise a method to build a dataset of images taken from a robot's perspective and we exploit it to obtain a door--status detector based on deep learning. We then leverage the typical working conditions of a robot to \emph{qualify} the model for boosting its performance in the working environment via fine--tuning with additional data. Our experimental analysis shows the effectiveness of this method with results obtained both in simulation and in the real--world, that also highlights a trade--off between the costs and benefits of the fine--tuning approach.

\end{abstract}

\section{Introduction}

Autonomous mobile robots are nowadays increasingly employed for cooperating with humans in a variety of tasks settled in indoor public, private, and industrial workplaces. A challenge posed to these \emph{service robots} is coping with highly dynamic environments characterised by features that can rapidly and frequently change, very often due to the presence of human beings~\cite{Kunze2018}. Consider, as examples, a domestic setup in an apartment or a workspace with several offices. In a time span of hours or days, the topology itself of these environments might frequently change its connectivity, since doors may be left open or closed, hence modifying in time the reachability of free spaces. This phenomenon strongly impacts the capability of robots to efficiently navigate and perform their tasks. At the same time, during their operational time, robots are often exposed to large amounts of data about their surroundings that offer an opportunity to track, model, and predict doors' statuses (and topology variations). The relevance of this problem is well--established in the literature. Different works, such as~\cite{longtermnavigation, dynamicmaps}, show how modelling the status of doors across a long time span and predicting the changes in the environment topology improves a robot's task performance. Intuitively, better paths can be planned by taking into account whether a room will be reachable or not upon arriving there.%

\begin{figure}[t]
    \centering
    \includegraphics[width=0.9\linewidth]{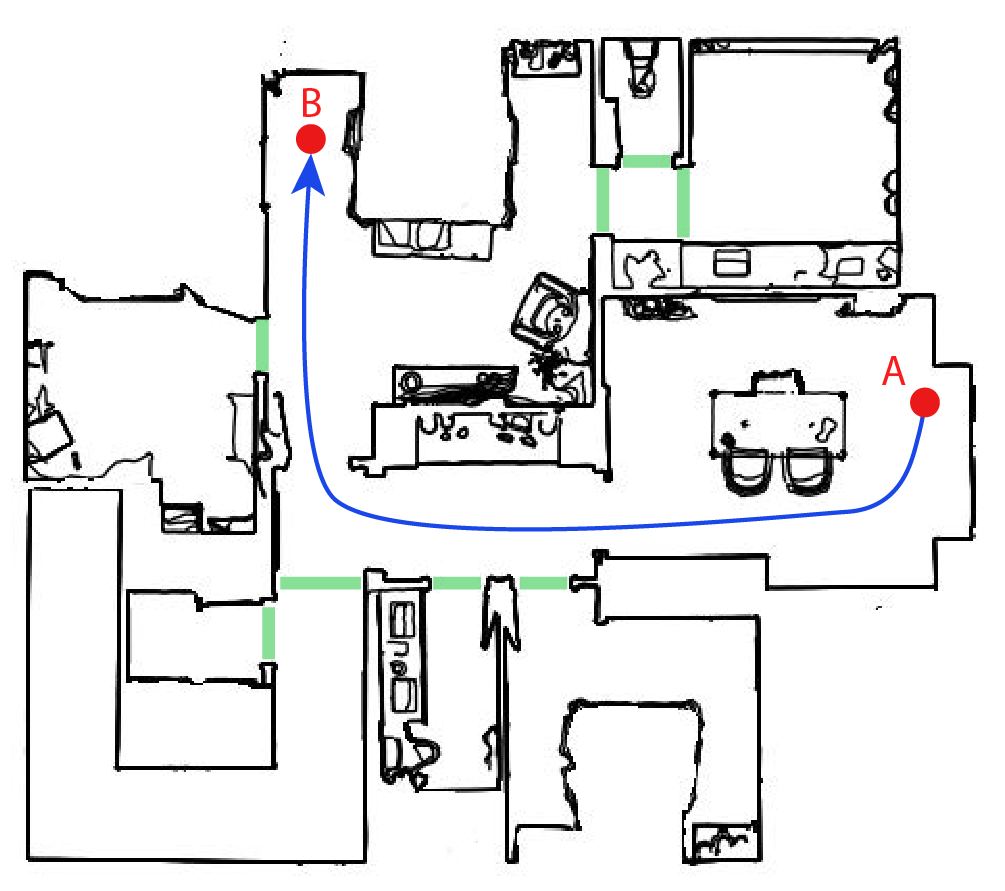}\\
    \subfloat[\label{fig:exa}]{\includegraphics[width=0.32\linewidth]{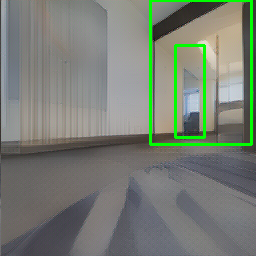}}
    \hfill
    \subfloat[\label{fig:exb}]{\includegraphics[width=0.32\linewidth]{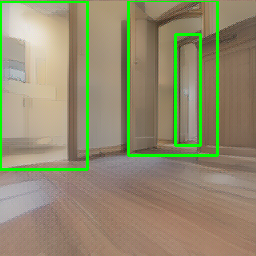}}
    \hfill
    \subfloat[\label{fig:exc}]{\includegraphics[width=0.32\linewidth]{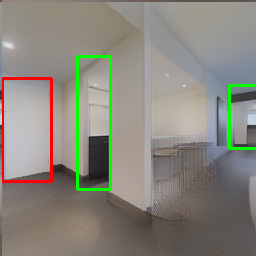}}
    \caption{A robot, navigating from A to B, can observe the status (open or closed) of different doors (highlighted in the top image). In this condition, door--status detection can be a difficult task as, from the robot's point of view, doors can be nested (\protect\subref*{fig:exa}--\protect\subref*{fig:exb}), doors can be hidden in the wall \protect\subref{fig:exc}, or instead of a door sometimes there are just passages (\protect\subref*{fig:exa}--\protect\subref*{fig:exc}). The bounding boxes of open (closed) doors, as identified by our method, are shown in green (red). For the remainder of this work, we follow the same colour schema.}
    \label{fig:door_examples}
\end{figure}

Central to unlocking such enhanced indoor navigation behaviours is what we call in this work \emph{door--status detection}: the robot's capability to extract, from visual perceptions, the presence and location of a door and, at the same time, to recognise its traversability (open or closed status).

In this work, we propose a method to endow a robot with door--status detection capabilities that can be run during task--related autonomous navigation.

Door--status detection is particularly challenging for mobile robots operating indoor since clear and well--framed views of a door are seldom encountered during navigation. 
Fig.~\ref{fig:door_examples} depicts some typical instances of these challenges. While navigating, the robot can view nested doors (Fig.~\ref{fig:exa}, \ref{fig:exb}), doors that are partially occluded (Fig.~\ref{fig:exb}, \ref{fig:exc}), or closed doors difficult to distinguish from their background (Fig.~\ref{fig:exc}).

To tackle the above problem, our approach starts with the choice of modelling door--status detection as a variant of object detection (OD) performed with deep neural networks. However, we found that OD deep learning methods, despite their great capabilities, exhibit important shortfalls when cast into the indoor robot navigation setting. Hence, our approach proposes a deployment methodology specific for mobile robots that allows harnessing the potentials of OD based on deep learning while solving what we recognised as the two most important limitations of such techniques in this domain.

First, OD methods are usually trained on large--scale datasets whose images are acquired from a human point of view. As a result, training examples follow a distribution that could be significantly different from the one generating the data perceived by a mobile robot. We show how popular datasets employed to train state--of--the--art deep learning detectors~\cite{coco, pascal}, do not properly represent the embodied perception constraints and uncertainty typically characterising a mobile robot~\cite{surveydeeplimits}, thus causing generalisation issues. 

Second, deep--learning OD modules are commonly trained with the main objective of obtaining a \emph{general detector}. This model is trained once, stored, and is meant to work in previously unseen environments. These practices are not optimal when considering the typical working conditions of an indoor service robot. After an initial deployment phase, the robot is commonly used in the same environment for a long time, sometimes even for its entire life cycle. In such persistent conditions, the robot eventually observes the same doors multiple times, from different points of view, and under various environmental conditions. Also, different doors may present similar visual features (e.g., multiple doors of the same model). Against this operative background, and from a practical point of view, the ability to generalise in new environments becomes less important, while correctly performing door--status detection in challenging images from the deployment environment becomes paramount.

To address the first limitation, we devise a method for acquiring a large visual dataset from multiple photorealistic simulations taking into account the robot's perception model along realistic navigation paths. This allows us to train a deep \emph{general} door--status detector with examples following a distribution compliant with the robot's perception capabilities.
To deal with the second limitation, we exploit the robot's operational conditions to tailor our general detector for a given target environment. 
We obtain what we call a \emph{qualified detector}, whose performance can substantially improve from the robot's experience enabling door--status detection in challenging instances (see the examples of Fig.~\ref{fig:door_examples}). Our solution relies on fine--tuning sessions~\cite{resnet, fasterrcnn, yolo} of the general detector (which shall be considered as a baseline) with new examples from the target environment. These data can be collected and labelled, for example, during the robot installation phases or while the robot carries out its duties. (A setting motivated also by our on--the--field experience with assistive robots~\cite{GIRAFF}.) %

We evaluate our approach by assessing its performance, also in the challenging cases exemplified in Fig.~\ref{fig:door_examples}, %
with an extensive experimental campaign conducted in simulated settings and in different real--world environments and conditions, as perceived by a mobile robot during its deployment. %

\section{Related Works}

Detecting a door's location can be useful for several tasks, as \emph{room segmentation}~\cite{segmentationsurvey}, i.e., to divide the map of the environment into semantically meaningful regions (rooms), to predict the shape of unobserved rooms~\cite{ECMR21}, or to do \emph{place categorisation}~\cite{scenerecognitiononjectdetection, placecategorization}, which assigns to the rooms identified within the occupancy map a semantic label (e.g., \emph{corridor} or \emph{office}) according to their aspect.

Recent studies~\cite{dynamicmaps,longtermnavigation} show how recognising door statuses can improve the navigation performance of robots in long--term scenarios. The work of~\cite{dynamicmaps} models the periodic environmental changes of a dynamic environment in a long--term run, while~\cite{longtermnavigation} proposes a navigation system for robots that operate for a long time in indoor environments with traversability changes.

Detecting doors in RGB images has been addressed as an OD task. Classical methods are based on the extraction of handcrafted features~\cite{humanoid, sonarandivisualdoordetection, edgeandcornerdoorsdetector}. %
Deep learning end--to--end methods~\cite{deeplearningoverview} provide significant improvements thanks to their capability of automatically learning how to characterise an object class, robustly  to scale, shift, rotation, and exposure changes. %
As a significant example, the work of~\cite{doorsandnavigation} describes a method for door detection with the goal of supporting and improving the autonomous navigation task performed by a mobile robot. A convolutional neural network is trained to detect doors in an indoor environment and its usage is shown to help a mobile robot to traverse passages in a more efficient way. Another approach, proposed in~\cite{doorcabinet}, focuses on robustly identifying doors, cabinets, and their respective handles in order to allow grasping by a robot. The authors use a deep architecture based on YOLO~\cite{yolo} to detect the Region Of Interest (ROI) of doors. This allows to obtain the handle's location by focusing only on the area inside the door ROI.

These works are representative examples of methods partially addressing the door--status detection problem in the mobile robotics domain. Indeed they do not explicitly consider the point of view of a mobile robot or do not take advantage of the robot's typical operational conditions. In this work, we devise an approach to overcome such limits.

\section{Building a Doors Dataset for Mobile Robots}
\label{sec:dataset}

One of the key prerequisites to exploit deep learning to synthesise an effective door--status detector for a mobile robot is the availability of a dataset consistent with its challenging 
perception model (see Fig.~\ref{fig:door_examples}).
The examples contained in the dataset should follow three main desiderata. Images (i) should represent different environments with different features, thus allowing the model to learn how to generalise; (ii) should contain doors as observed from a point of view similar to the one of a robot navigating in an indoor environment; (iii) should be taken from real environments or with an adequate level of photorealism.

An effective but impractical and time--consuming way to comply with the above requirements would be to deploy a robot on the field and having it exploring different environments while acquiring image samples of doors.
The large overheads of such a procedure are well--known and a popular alternative is to rely on simulations~\cite{collins2021review} or publicly available datasets~\cite{coco, imagenet, deepdoors2}. 

Meeting the desiderata (i)-(iii) in simulation is not straightforward since these are seldom guaranteed by available frameworks.
For example, simulation tools popular in robotics such as Gazebo~\cite{koenig2004design} or Unreal~\cite{USARSIM}, while providing accurate physics modelling, fail to represent the realism and complexity of the perceptions in the real world. At the same time, public datasets as~\cite{coco,imagenet,deepdoors2}, do not well represent the point of view of a robot in its working conditions~\cite{surveydeeplimits}. To address these issues, we resorted to Gibson~\cite{gibson}, a simulator for embodied agents that focuses on realistic visual perceptions, and to the environments from Matterport3D~\cite{matterport}, an RGB--D dataset of 90 real--world scans.
Given a simulated environment, we extract a set of poses that could describe views compatible with a mobile robot by applying a set of principles; the key ones include lying in the reachable free space (feasibility), ensuring a minimum clearance from obstacles, and being along the shortest paths between key connecting locations in the environment's topology.
We achieve them with an extraction algorithm working in three phases: grid extraction, navigation graph extraction, and pose sampling.

The grid extraction phase aims at obtaining a 2D occupancy grid map, similar to those commonly used by mobile robots for navigation. We start from the environment's 3D mesh, and we aggregate obstacles from multiple cross--sections of the 3D mesh performed with parallel planes. %
The result is then manually checked for inaccuracies and artefacts produced during the procedure. %

\begin{figure}[!tbp]
	\centering
    \subfloat[Navigation graph\label{fig:pose_estimator_pruned_lines}]{\includegraphics[width=0.5\linewidth]{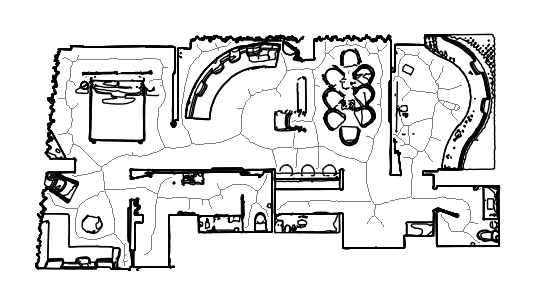}}
    \subfloat[Sampled poses\label{fig:pose_estimator_subsampled}]{\includegraphics[width=0.5\linewidth]{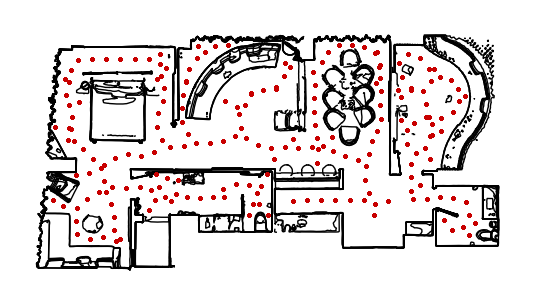}}
	\caption{Different phases of the pose extraction algorithm.}
	\label{fig:pose_estimator}
\end{figure}

The \emph{navigation graph} (shown in Fig.~\ref{fig:pose_estimator_pruned_lines}) is a data structure that we use to represent the topology of the locations on the grid map that correspond to typical waypoints a robot occupies while navigating in the environment. We compute it from a Voronoi tessellation of the grid map by using obstacle cells as basis points~\cite{thrun1996integrating}, extracting graph edges from those locations that maintain maximum clearance from obstacles. 
%

We then perform pose sampling on the navigation graph. The algorithm extracts from the graph a list of positions keeping a minimum distance $D$ between them (this parameter controls the number and the granularity of the samples). A visual example of the algorithm's results is shown in Fig.~\ref{fig:pose_estimator_subsampled}. 

To build the dataset we acquire an image from the points of view of a robot's front--facing camera simulating its perceptions in the virtual environment from the sampled poses. %
Specifically, in each pose on the grid map, we acquire perceptions at two different height values (\SI{0.1}{\meter} and \SI{0.7}{\meter} -- to simulate different embodiments of the robots) and at $8$ different orientations (from $0^{\circ}$ to $315^{\circ}$ with a step of $45^{\circ}$). Each acquisition includes the RGB image, the depth information, and the semantic data from Matterport3D. %
Since the semantic annotation of Matterport3D presents some inaccuracies, data labelling is manually performed
by a human operator who specifies the door bounding boxes and the door status as open or closed. We considered $10$ different Matterport3D environments (small  apartments or large villas  with multiple floors and a heterogeneous furniture style) by setting $D = 1 \text{m}$. The final dataset we obtained is composed of approx. $5500$ examples.

\section{Door--Status Detection for Mobile Robots}\label{sec:models}

In this section, we first detail how we synthesised a \emph{General Detector} ($GD$, Section~\ref{sec:general_door_detector}) using a dataset generated with the approach of Section~\ref{sec:dataset}. Subsequently, leveraging the assumption that the environment $e$ will not change in its core features (location and visual aspect of doors) during the robot's long--term deployment, we introduce our \emph{Qualified Detector} for $e$ ($QD_e$) by applying a procedure based on fine--tuning~\cite{resnet,fasterrcnn,yolo,detr} of the $GD$ on additional data that, in our envisioned scenario, can be acquired and labelled during the first setup of the robot in $e$.

%

%

%

%

%

%

\subsection{General Door--Status Detector}
\label{sec:general_door_detector}

As previously introduced, we aim at building and deploying door--status detectors for mobile robots leveraging deep--learning for object detection. In a preliminary experimental phase, we evaluated and compared three popular models suitable for such a task: DETR~\cite{detr}, Faster--RCNN~\cite{fasterrcnn}, and a YOLO architecture~\cite{yolov3}. We decided to adopt DETR since, with respect to the other two methods, it turned out to be easier to deploy in our robotic setting primarily due to two key features.
First, DETR does not require setting in advance the number and dimension of anchors (i.e., sets of predefined bounding boxes used to make detections) according to the image resolution and the objects' shape, a task that instead the YOLO architecture requires. Second, both competitors require a final non--maximum suppression step to discard multiple detections of the same object. DETR, instead, matches each bounding box to a different object by construction. Hence, the methodology we describe in this paper and the empirical results evaluating it shall develop around DETR--based detectors. However, we stress the fact that our methods can be applied to any architecture, including those mentioned above, and, eventually, to their improvements.

DETR combines a CNN backbone based on ResNet~\cite{resnet} to produce a compact representation of an image and a transformer~\cite{transformer}. %
We used the pre--trained version of DETR on the COCO~2017~\cite{coco} dataset and, to adjust for door--status detection, we chose the smallest configuration provided by the authors. %

The model requires setting one hyper--parameter, $N$, which determines the fixed number of bounding boxes predicted for each image. As a consequence, to filter out the detected doors, we select the $n \leq N$ bounding boxes whose confidence is not below a threshold $\rho_{c}$. We tuned $N$ to be higher (but close) to the maximum number of doors in any single image of our dataset.
To train the general detector, we fixed the first two layers of the CNN backbone (as in~\cite{detr}) with the weights of the pre--trained model. We then re--trained the remaining layers with images from the dataset of Section~\ref{sec:dataset}. To achieve data augmentation, we generated additional samples by applying a random horizontal flip and resize transformation to a subset of the images (each training sample is selected for this procedure with a probability of $0.5$).

\subsection{Qualification on a Target Environment}\label{sec:qualified_detector}

Given a new environment $e$ we use a randomly sampled subset of the images collected in it to fine--tune the $GD$, obtaining the qualified detector $QD_e$. To be used in the fine--tuning procedure, these images need to be labelled specifying the bounding boxes and the status for each visible door.  

In our envisioned scenario, this data acquisition and labelling tasks can be carried out by a technician during the robot's first installation in $e$ or in a second phase by uploading the data to a remote server. Such a setup phase requires to build the map of the environment (either autonomously or with teleoperation) by observing the entirety of the working environment and is very relevant to many real--world installations of collaborative robots, as we recently experienced with extensive on--the--field testing in the use case of assistive robotics~\cite{GIRAFF}.
This manual labelling task is quite time--expensive; yet it is required. In principle, we could use \emph{pseudo--labels} automatically obtained by running the $GD$ over the additional samples for incremental learning. Despite intriguing, we empirically observed that this is particularly challenging due to the fact that pseudo--labels are not enough accurate for this process. Recently, the work of \cite{pseudolabelsRAL} showed how pseudo--labels are particularly noisy and inaccurate: while they can be used to improve performance in tasks where precise labels are less important (like semantic segmentation), they are still too inaccurate to be used in object detection tasks, like the one investigated in this work. We observed how fine--tuning a general detector using pseudo--labels results in a performance degradation of about $20\%$ when compared with the $GD$. These challenges are well-known and the approach we follow in this work is customary. See, for example, the work of~\cite{localizationobjectdetectionwithpriormap}, where manual annotations have been used to label 3D objects to fine--tune a model employed in long--term localisation. The study of methods to ease the burden of this task will be addressed as a part of our future work.

Finally, note that, while we focus here on a specific $GD$ based on DETR~\cite{detr}, this method to obtain a qualified detector $QD$ is general and can be applied to other deep learning--based models, such as YOLO~\cite{yolov3} or Faster--RCNN~\cite{fasterrcnn}.

\section{Evaluation in Simulation}

\subsection{Experimental Setting}
\label{sec:exp_setting_sim}

We evaluate our method using simulated data $\mathcal{D}$ obtained, as described in Section~\ref{sec:dataset}, from $10$ different Matterport3D~\cite{matterport} environments. We test the performance of our detectors on each environment $e$ independently. First, we train the general detector $GD_{-e}$ using the dataset $\mathcal{D}_{-e}$, where $\mathcal{D}=\{\mathcal{D}_{-e}, \mathcal{D}_e\}$,  $\mathcal{D}_{e}$ contains all the instances acquired from poses sampled in environment $e$, and $\mathcal{D}_{-e} = \mathcal{D} \setminus \mathcal{D}_e$. This general detector will be used as a baseline in most of the evaluations we present. Then, we randomly partition the first subset as $\mathcal{D}_{e} = \{\mathcal{D}_{e, 1}, \mathcal{D}_{e, 2}, \mathcal{D}_{e, 3}, \mathcal{D}_{e, 4}\}$, where each $\mathcal{D}_{e,i}$ contains the $25\%$ of the examples from $e$, randomly selected. 

While $\mathcal{D}_{e, 4}$ is reserved for testing, the remaining subsets are used to perform a series of fine--tuning rounds to obtain the corresponding qualified door--status detectors. Specifically, we fine--tune $GD_{-e}$ using these three additional data subsets: $\{\mathcal{D}_{e, 1}\}$, $\{\mathcal{D}_{e, 1},\mathcal{D}_{e, 2}\}$, and $\{\mathcal{D}_{e, 1},\mathcal{D}_{e, 2},\mathcal{D}_{e, 3}\}$. We denote the obtained qualified detectors as $QD^{25}_{e}$, $QD^{50}_{e}$, and $QD^{75}_{e}$, respectively. The superscript denotes the percentage of data instances from $e$ that are required to fine--tune the general door--status detector. Such a percentage can be interpreted as an indicator of the cost to acquire and label the examples. To give a rough idea, labelling the $25\%$ of the dataset (approximately $150$ images) took a single human operator an effort of about $1$ hour.

We empirically set the parameters of the door--status detector as $N = 10$ (number of bounding boxes) and $\rho_{c} = 0.75$ (confidence threshold). We conducted an extensive preliminary experimental campaign spanning different batch sizes ($\{1,2,4,16,32\}$) and the number of epochs ($\{20,40,60\}$) selecting $1$ and $60$ for the general detector and $1$ and $40$ for the qualified ones, respectively.

We measure performance with the average precision score (AP) used in the Pascal~VOC challenge~\cite{pascal} by adjusting for a finer interpolation of the precision/recall curve to get a more conservative (in the pessimistic sense) evaluation. The AP is a popular evaluation metric widely adopted for object detection tasks, it represents the shape of the precision/recall curve as the mean precision 
over evenly distributed levels of recall. To accept a true positive, the bounding box computed by the network must exhibit an Intersection Over Union area ($IOU$) with one true bounding box above a threshold $\rho_{a}$, that we empirically set to $50\%$.

The source code of our simulation framework (Section~\ref{sec:dataset}),
the door--status detectors (Section~\ref{sec:models}), and the collected datasets are
maintained in a freely accessible repository\footnote{ \url{https://github.com/aislabunimi/door-detection-long-term}}.

\subsection{Results}

\begin{figure}[!htbp]
	\centering
        \includegraphics[width=0.91\linewidth]{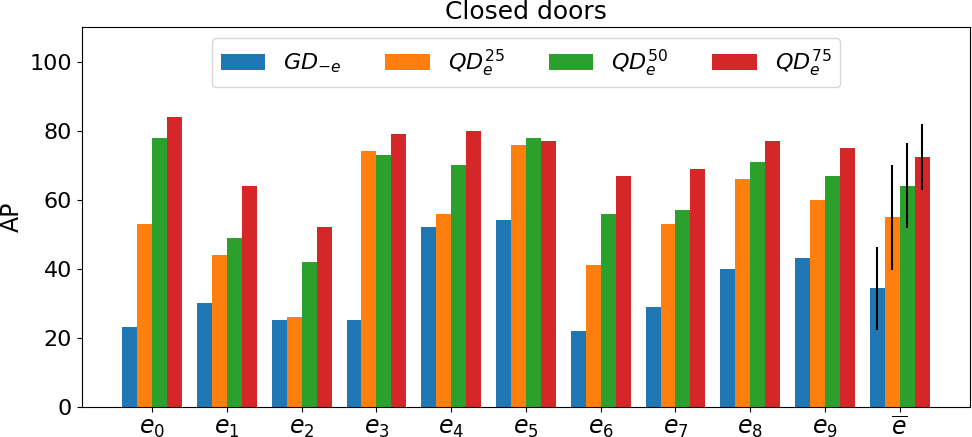} \\\vspace{0.3em}
        \includegraphics[width=0.91\linewidth]{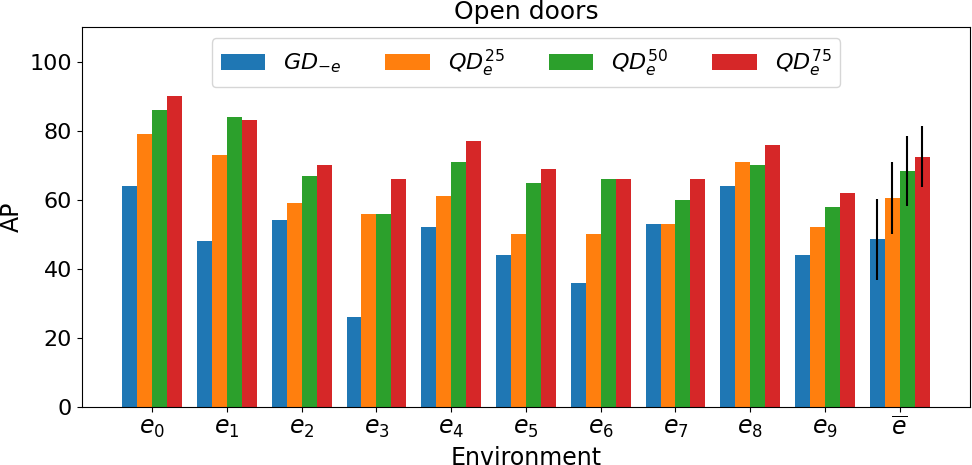}
	\caption{AP scores in Matterport3D environments.}
	\label{fig:all_houses}
\end{figure}

Table~\ref{tab:all_houses_results} reports the mean AP scores (averaged over the $10$ environments) reached by the $4$ detectors divided by label (closed door, and open door), the average increments (with respect to the detector immediately above in table) obtained with fine--tuning, and the standard deviation ($\sigma$). We also report the AP scores for every environment in Fig.~\ref{fig:all_houses}. These results show the trade-off between performance increase (via fine--tuning) and costs due to data collection and labelling.

\begin{table}[!htbp]
	\centering
	\begin{scriptsize}
	\begin{tabular}{cccccc}
		\toprule
		 \textbf{Exp.} & \textbf{Label} & \textbf{AP} & \textbf{$\sigma$} & \textbf{Increment} &  \textbf{$\sigma$} \\
		\midrule
		\multicolumn{1}{c|}{\multirow{2}{*}{$GD_{-e}$}} & Closed & 34 & 12 &  -- & -- \tabularnewline 
\multicolumn{1}{c|}{} & Open  & 48 & 12 &  -- & -- \\  \hline 
\multicolumn{1}{c|}{\multirow{2}{*}{$QD^{25}_e$}} & Closed & 55 & 15 & $70\%$ & 58\tabularnewline 
\multicolumn{1}{c|}{} & Open  & 60 & 10 & $30\%$ & 34\\  \hline 
\multicolumn{1}{c|}{\multirow{2}{*}{$QD^{50}_e$}} & Closed & 64 & 12 & $21\%$ & 21\tabularnewline 
\multicolumn{1}{c|}{} & Open  & 68 & 10 & $14\%$ & 11\\  \hline 
\multicolumn{1}{c|}{\multirow{2}{*}{$QD^{75}_e$}} & Closed & 72 & 10 & $14\%$ & 9\tabularnewline 
\multicolumn{1}{c|}{} & Open  & 72 & 9 & $7\%$ & 5\\
		\bottomrule
	\end{tabular}
	\end{scriptsize}
	\caption{Average AP in Matterport3D environments.}
	\label{tab:all_houses_results}
\end{table}

\begin{figure}[!htbp]
	\centering
    \includegraphics[width=0.32\linewidth]{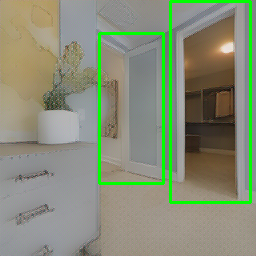}
	\hfil
	\includegraphics[width=0.32\linewidth]{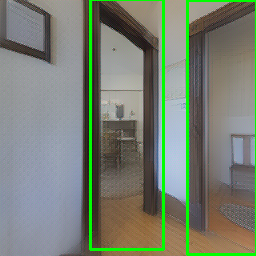}
	\hfil
	\includegraphics[width=0.32\linewidth]{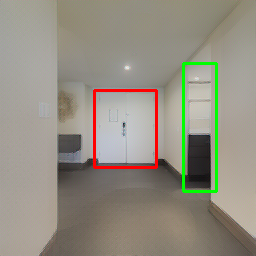}
	\caption{Door--statuses found by the general detector $GD_{-e}$.}
	\label{fig:gd_examples}
\end{figure}

Results from Table~\ref{tab:all_houses_results} and Fig.~\ref{fig:all_houses} show that the general detector $GD_{-e}$, thanks to our dataset's consistency with the robot's perception model (see Section~\ref{sec:dataset}), is able to correctly detect doors statuses in those cases where they are clearly visible, as shown in Fig.~\ref{fig:gd_examples}. However, while such a performance allows its use on a robot, there is significant room for improvement in detecting more challenging examples.

\begin{figure}[!htbp]
	\centering
    \includegraphics[width=0.32\linewidth]{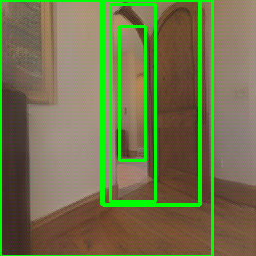}
	\hfill
    \includegraphics[width=0.32\linewidth]{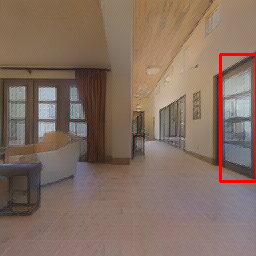}
    \hfill
    \includegraphics[width=0.32\linewidth]{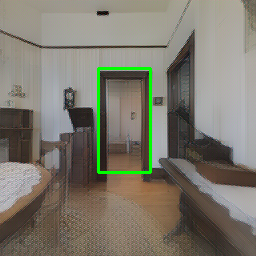}
    \\\vspace{0.15cm}
	\includegraphics[width=0.32\linewidth]{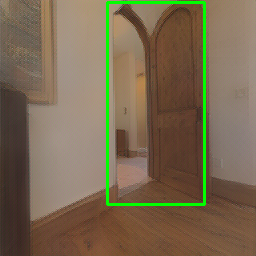}
	\hfill
	\includegraphics[width=0.32\linewidth]{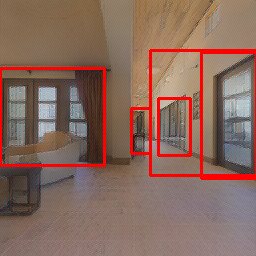}
	\hfill
	\includegraphics[width=0.32\linewidth]{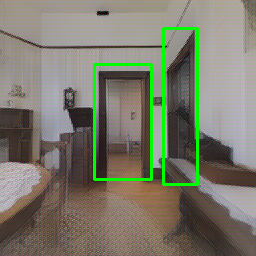}
	\caption{Door--statuses as identified by $GD_{-e}$ (top row) compared to $QD^{25}_{e}$ (bottom row) in Matterport3D environments. }
	\label{fig:qd_25_examples}
\end{figure}

More interestingly, qualified detectors achieve a steep increase in performance. Unsurprisingly, the performance improves with more data (and data preparation costs) from $QD^{25}_{e}$ to $QD^{75}_{e}$. However, it can be seen how $QD^{25}_{e}$, despite requiring a relatively affordable effort in manual labelling, obtains the highest performance increase. From a practical perspective, this shows how the availability of \emph{a few} labelled examples from the robot target environment could be a good compromise between performance and costs to develop an environment--specific door--status detector. This suggests that the number of examples that have to be collected and labelled on the field can be limited, thus promoting the applicability of our proposed framework. An example of this is shown in Fig.~\ref{fig:qd_25_examples}, where it can be seen how a $QD^{25}_{e}$ (bottom row) fixes the mistakes of its corresponding general detector $GD_{-e}$ (top row) in challenging images with nested or partially observed doors.

\section{Evaluation in the Real World}

\subsection{Experimental Settings}
\label{sec:exp_setting_real}
In this section, we evaluate the performance of our method with a real robot by using images collected by a \mbox{Giraff--X} platform~\cite{GIRAFF} (Fig.~\ref{fig:giraff}) during a teleoperated exploration of $3$ single--floor indoor environments with multiple rooms from two buildings in our campus. Images were extracted from the robot's perceptions during navigation at $1$ fps.
As it commonly happens with real--world robot data, images are acquired in noisy environmental conditions with low--quality cameras, thus making the detection task even more difficult. In our setting, we used 320x240 RGB images acquired with an Orbbec Astra RGB--D camera.

First, we consider two general detectors. One is trained with simulated data $\mathcal{D}$, as in the previous section but with all the $10$ environments. Another one is trained with real--world images from the publicly available \emph{DeepDoors2} dataset (DD2)~\cite{deepdoors2}, which features $3000$ images of doors that we re-labelled to include the ground truth for challenging examples not originally provided. Comparing these two general detectors, we aim at assessing the advantages of training with a dataset following the principles we proposed in Section~\ref{sec:dataset} instead of relying on mainstream datasets for classical object detection.

Subsequently, following the same steps of Section~\ref{sec:exp_setting_sim}, we qualify both $GD$s by using the $25$, $50$, and $75\%$ of data collected in the three real environments.

\subsection{Evaluation metrics}
\label{sec:metrics_real}

Analogously to what is done in simulation, we report the AP scores, but we argue that the real--world evaluation of our method can be conducted also with additional metrics, which are more representative of the actual application domain where door--status detection is meant to be cast.

\begin{figure}[!htbp]
    \hspace*{\fill}
    \begin{minipage}{0.39\linewidth}
          \subfloat[\label{fig:giraff}]{\includegraphics[width=\linewidth]{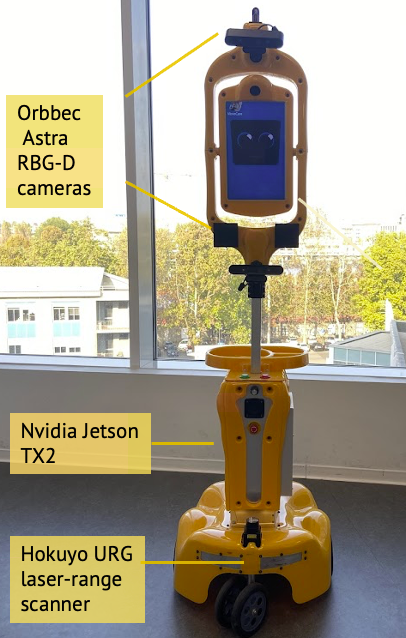}}
    \end{minipage}
    \hspace{0.1cm}
    \begin{minipage}{0.37\linewidth}
          \subfloat[\label{fig:ex_wrong_label}]{\includegraphics[width=\linewidth]{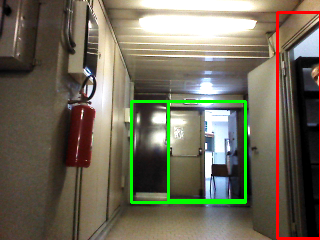}}\\\vspace{-0.3cm}
        \subfloat[\label{fig:ex_metric_bbox_localization}]{\includegraphics[width=\linewidth]{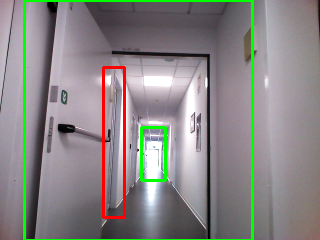}}
    \end{minipage}
    \hspace*{\fill}%
    \caption{Examples (\protect\subref*{fig:ex_wrong_label}--\protect\subref*{fig:ex_metric_bbox_localization}) of different types of errors made by a door--status detector mounted on our Giraff--X robot \protect\subref{fig:giraff}. While the AP considers all these errors in a similar way, our proposed metric considers them differently, according to their potential impact on robot performance.}
    \label{fig:metric_examples}
\end{figure}

The AP (as well as other metrics used in computer vision) presents some limitations when used in our context. Such a metric considers as false positives multiple bounding boxes assigned to the same door, as in Fig.~\ref{fig:ex_wrong_label}. %
However, a robot can easily disambiguate this by leveraging additional data such as its estimated pose and the map of the environment. On one side, although an erroneous localisation of the bounding box of a door (Fig.~\ref{fig:ex_metric_bbox_localization})  penalises the AP, it might have little effect in practice. On the other side, the AP is very marginally affected by a wrong detection of the door status if the bounding box is sufficiently accurate due to the fact that different labels are treated as two independent object classes, as the case in Fig.~\ref{fig:ex_wrong_label}. Conversely, the error of misleading a closed passage for an open one (and \emph{vice versa}) can significantly impact the robot's performance when the robot translates such information into actions.

To address this shortcoming and better capture performance in our robotic setting, we introduce three additional metrics. Consider a door $i$ in a given image. If multiple bounding boxes are matched to $i$, where matching means $IOU \ge \rho_a$, the one with the maximum above--threshold ($\rho_c$) confidence is selected. If the status of the door is correctly identified, we consider it as a true positive ($TP$). Otherwise, we classify it as a False Positive ($FP$). All the remaining bounding boxes matched to $i$ are, as per our previous considerations, ruled out from the evaluation. Finally, when a bounding box does not meet the $IOU$ condition with any door in the image, we count it as a Background False Detection ($BFD$). A False Positive and a Background False Detection are errors that can play very different roles inside a robotic use case. While the first is likely to affect the robot's decisions, the second one might increase the uncertainty in the robot world--model. We scale the above metrics using the true number of doors in the testing set, denoted as $GT$, thus obtaining $TP_{\%}=TP/GT$, $FP_{\%}=FP/GT$, and $BFD_{\%}=BFD/GT$.

\subsection{Results}

Table~\ref{tab:all_results_real_world} compares the average AP of the $GD$ trained with DD2~\cite{deepdoors2} with that trained with our dataset $\mathcal{D}$. Intuitively, a model trained with real--world data (such as those featured in DD2) should have higher performance when used with real--world images, if compared with a model trained with simulated data (as $\mathcal{D}$). However, Table~\ref{tab:all_results_real_world} shows how the $GD$ and $QD$s trained with $\mathcal{D}$ have higher performance than those trained with DD2. This is because training images of $\mathcal{D}$, collected from the simulated point of view of a robot, better represent the actual distribution of robot perceptions, allowing us to fill, to some extent, the sim--to--real gap. Moreover, Table~\ref{tab:all_results_real_world} shows that the fine--tuning operation to qualify general detectors to the target environment works remarkably well also when used in real--world conditions. For these reasons, from now on, we present results referring to the general detector trained with $\mathcal{D}$.

\begin{table}[!htbp]
\centering
\begin{scriptsize}
\begin{tabular}{p{0.07\linewidth}c|cccc|cccc}
\toprule
                  &  & \multicolumn{4}{c|}{\textbf{DeepDoors2 (DD2)} }      & \multicolumn{4}{c}{\textbf{Simulation dataset ($\mathcal{D}$)}} \\
                  \textbf{Exp.} & \textbf{Label} &\textbf{AP} & \textbf{$\sigma$} & \textbf{Inc.} &  \textbf{$\sigma$}  &   \textbf{AP} & \textbf{$\sigma$} & \textbf{Inc.} &  \textbf{$\sigma$}    \\
\midrule
\multirow{2}{*}{$GD$} & Closed & 5 & 3 &  -- & -- & 13 & 10 &  -- & -- \tabularnewline 
 & Open  & 18 & 5 &  -- & -- & 31 & 11 &  -- & -- \\  \hline 
\multirow{2}{*}{$QD^{25}_e$} & Closed & 33 & 9 & $631\%$ & 240& 53 & 9 & $508\%$ & 424\tabularnewline 
 & Open  & 43 & 14 & $134\%$ & 20& 55 & 14 & $83\%$ & 19\\  \hline 
\multirow{2}{*}{$QD^{50}_e$} & Closed & 52 & 7 & $66\%$ & 51& 65 & 8 & $24\%$ & 15\tabularnewline 
 & Open  & 51 & 14 & $18\%$ & 7& 70 & 7 & $29\%$ & 22\\  \hline 
\multirow{2}{*}{$QD^{75}_e$} & Closed & 55 & 8 & $5\%$ & 6& 72 & 8 & $10\%$ & 5\tabularnewline 
 & Open  & 65 & 8 & $32\%$ & 18& 78 & 8 & $13\%$ & 1\\
\bottomrule
\end{tabular}
\end{scriptsize}
\caption{Average AP in real--world environments when DD2 dataset and our one ($\mathcal{D}$) are used to train the $GD$.}
	\label{tab:all_results_real_world}
\end{table}

\begin{figure}[!htbp]
	\centering
    \includegraphics[width=0.32\linewidth]{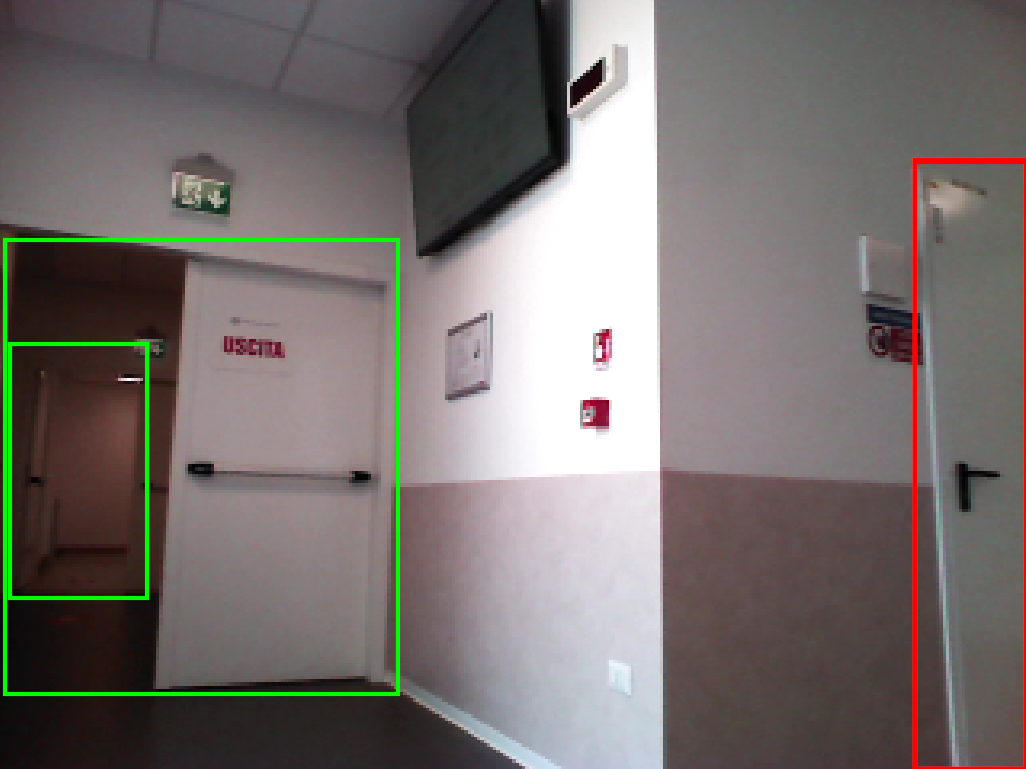}
	\hfill
    \includegraphics[width=0.32\linewidth]{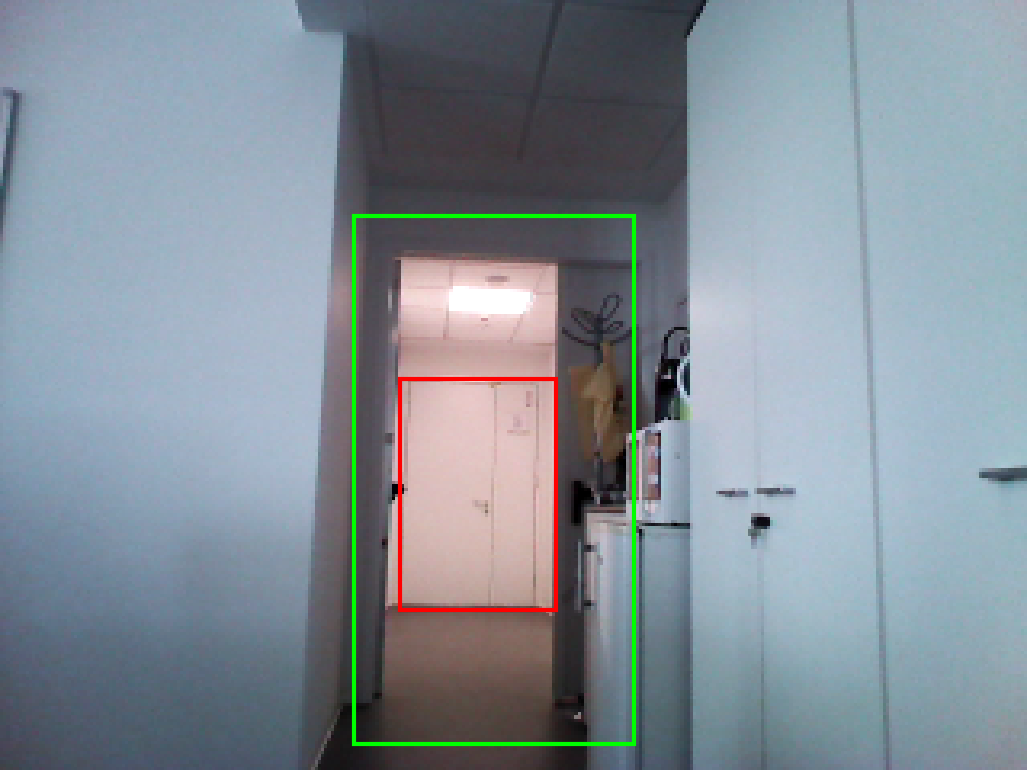}
    \hfill
    \includegraphics[width=0.32\linewidth]{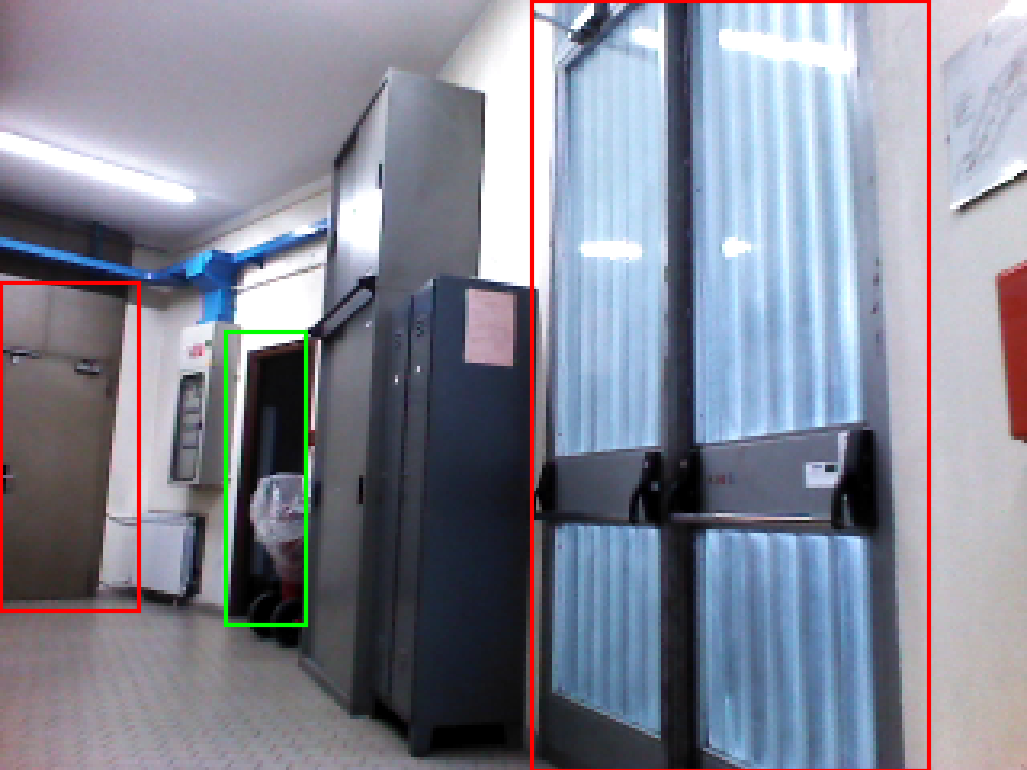}
    \\\vspace{0.15cm}
	\includegraphics[width=0.32\linewidth]{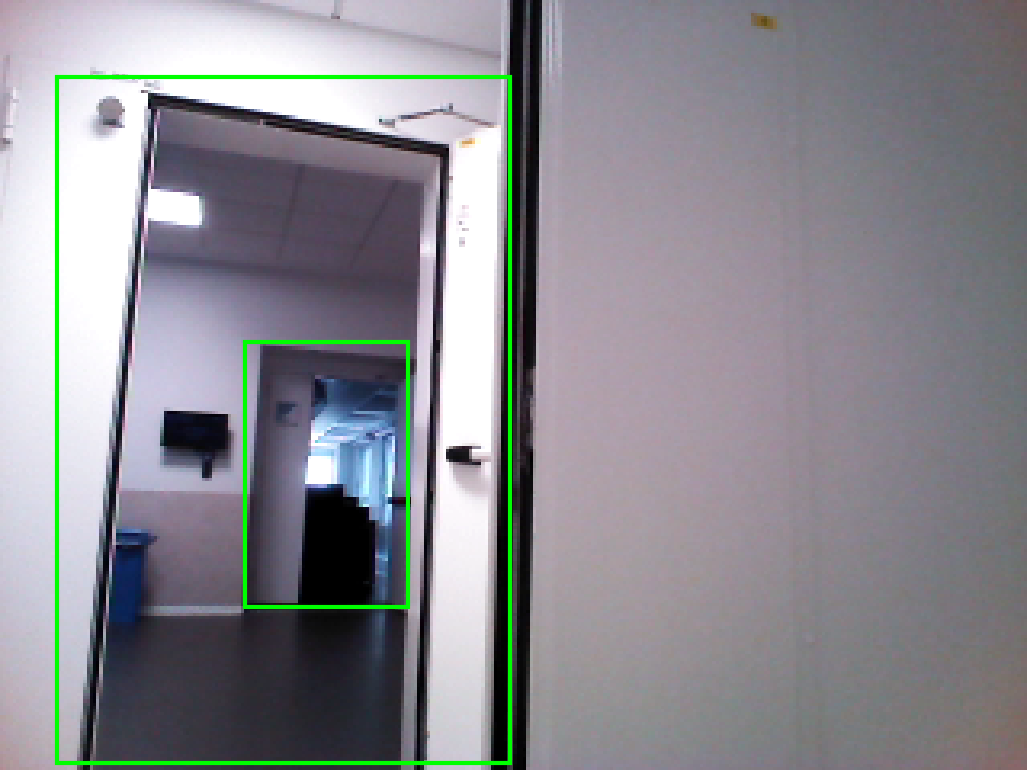}
	\hfill
	\includegraphics[width=0.32\linewidth]{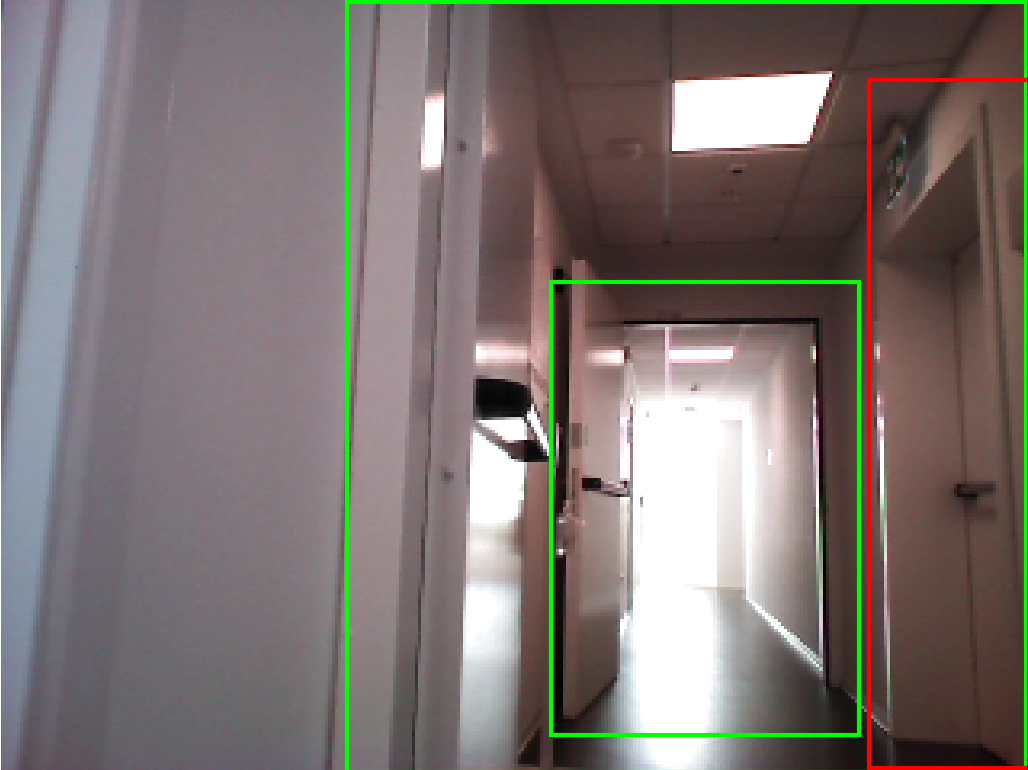}
	\hfill
	\includegraphics[width=0.32\linewidth]{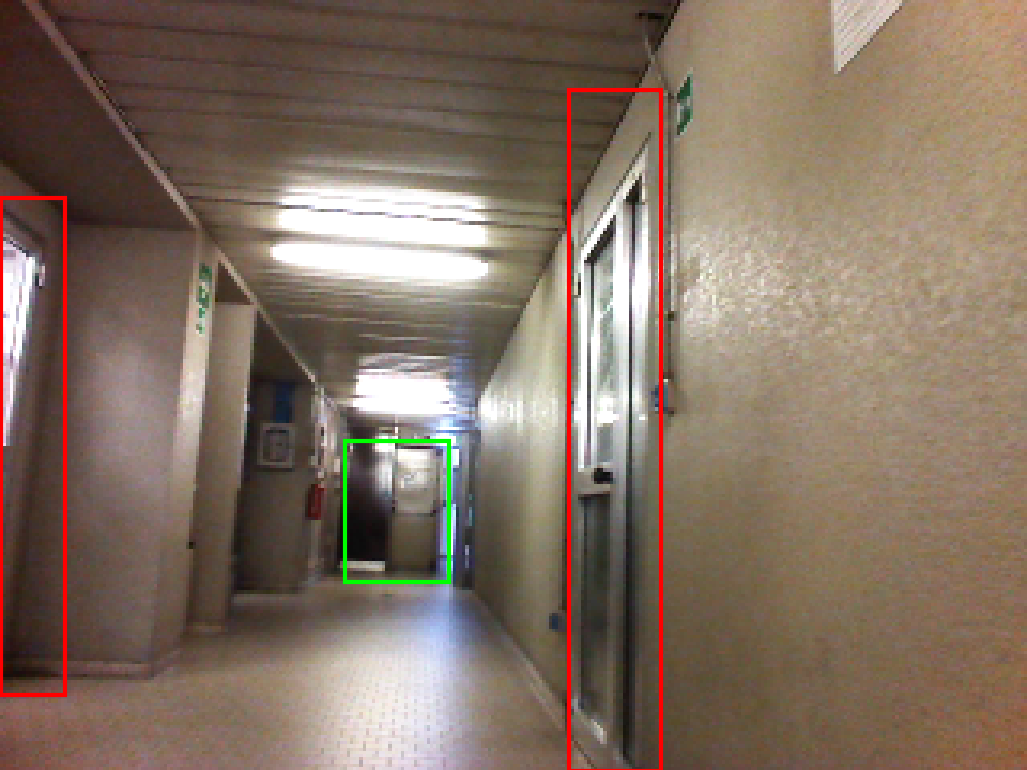}
	\caption{Challenging door--statuses detected by $QD^{25}_e$ in the real environments $e_1$, $e_2$, and $e_3$ (ordered by columns).}
	\label{fig:qd_25_real_examplez}
\end{figure}

The performance of $QD$s is similar to that obtained in the far less challenging dataset of Table~\ref{tab:all_houses_results}. Most importantly, the performances of $QD^{25}_{e}$ corroborate our findings from Section~\ref{sec:exp_setting_sim}, confirming how few additional examples can induce a significant increase in performance. Beyond the improvements observed in the average scores, $QD^{25}_{e}$ managed to provide correct door--status detection in very challenging cases. We report in Fig.~\ref{fig:qd_25_real_examplez} some representative examples. As it can be seen, our detectors correctly recognised cases with nested doors, partially visible frames, and narrow side views.
Another relevant example is in the second image (top row) of Fig.~\ref{fig:qd_25_real_examplez}, where the qualified model succeeds in detecting a white closed door in the background while, at the same time, not making a false detection of the white wardrobe doors on the right.

\begin{table}[!htbp]
\centering
\begin{scriptsize}
\begin{tabular}{c|ccccc}
\toprule
\textbf{Env.} & \textbf{Exp.} & $GT$ & $TP$ ($TP_{\%}$) & $FP$  ($FP_{\%}$) & $BFD$ ($BFD_{\%}$)  \\
\midrule
\multirow{4}{*}{$e_1$} &$GD$ & 235 & 71 (30\%) & 18 (7\%) & 51 (21\%) \\ 
 & $QD^{25}_e$ & 235 & 145 (61\%) & 10 (4\%) & 64 (27\%) \\ 
 & $QD^{50}_e$ & 235 & 179 (76\%) & 4 (1\%) & 44 (18\%) \\ 
 & $QD^{75}_e$ & 235 & 190 (80\%) & 4 (1\%) & 36 (15\%) \\ 
[2pt]\hline 
\multirow{4}{*}{$e_2$} &$GD$ & 269 & 96 (35\%) & 17 (6\%) & 56 (20\%) \\ 
 & $QD^{25}_e$ & 269 & 192 (71\%) & 11 (4\%) & 87 (32\%) \\ 
 & $QD^{50}_e$ & 269 & 206 (76\%) & 6 (2\%) & 66 (24\%) \\ 
 & $QD^{75}_e$ & 269 & 228 (84\%) & 7 (2\%) & 60 (22\%) \\ 
[2pt]\hline 
\multirow{4}{*}{$e_3$} &$GD$ & 327 & 62 (18\%) & 19 (5\%) & 108 (33\%) \\ 
 & $QD^{25}_e$ & 327 & 183 (55\%) & 22 (6\%) & 190 (58\%) \\ 
 & $QD^{50}_e$ & 327 & 230 (70\%) & 13 (3\%) & 103 (31\%) \\ 
 & $QD^{75}_e$ & 327 & 248 (75\%) & 8 (2\%) & 75 (22\%) \\ 
 \bottomrule
\end{tabular}
\end{scriptsize}
\caption{Extended results in the real--world environments.}
\label{tab:metric_complete}
\end{table}

Table~\ref{tab:metric_complete} reports the detailed results, for all three environments, of the metrics we defined in Section~\ref{sec:metrics_real}. 
The results show that $GD$, although it has a low number of wrong predictions ($FP$ and $BFD$), is capable of detecting only a few of the $GT$ doors in the images ($TP$). On the contrary, $QD$s dramatically improve performance, with $QD^{25}_{e}$ showing a $TP_{\%}$ of $62\%$ on average.

\begin{figure}[!htbp]
	\centering
	\includegraphics[width=0.95\linewidth]{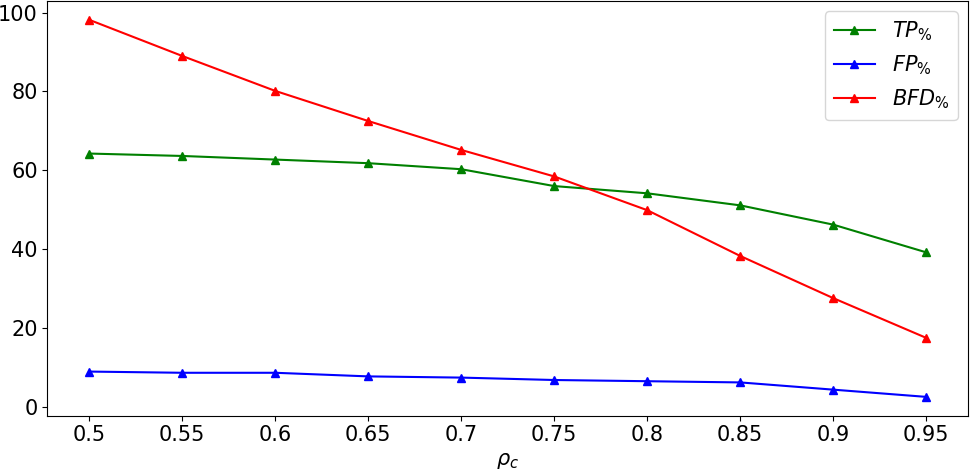}
	\caption{$TP_{\%}$, $FP_{\%}$, and $BFD_{\%}$ obtained by $QD^{25}_{e_3}$ for increasing confidence thresholds.}
	\label{fig:confidence_study}
\end{figure}

Among the three environments considered, we argue that $e_3$ is the more challenging, as it can be seen by the higher number of $BFD$. To cope with this, there are two possible directions. First, increasing the number of manually labelled examples reduces $BFD$ (as can be seen already with $QD^{50}_{e}$). Alternatively, adopting a more conservative selection rule by increasing the confidence threshold $\rho_c$, at the cost of slightly reducing the number of $TP$. In Fig.~\ref{fig:confidence_study}, we show how $TP_{\%}$, $FP_{\%}$, and $BFD_{\%}$ for $QD^{25}_{e_3}$ change when varying $\rho_c$ in $e_3$. Such an instance confirms how $\rho_c=0.75$ is an acceptable trade--off among $TP_{\%}$ (high) and $BFD_{\%}$ (low) for such a detector.

In long--term runs, the illumination conditions of an environment might change from those of the initial setup, and this may affect the performance of the door--status detector. To test the robustness of our approach to this event, we acquire (following the same procedure of Section~\ref{sec:exp_setting_real}) data from environments $e_1$ and $e_2$ during nighttime, when only artificial light is present and some rooms are dark. Then, we use these images to test the $GD$ and the $QDs$ fine-tuned with data acquired during the initial setup time, with daylight.

\begin{table}[!htbp]
	\centering
	\begin{scriptsize}
	\begin{tabular}{cccccc}
		\toprule
		 \textbf{Exp.} & \textbf{Label} & \textbf{AP} & \textbf{$\sigma$} & \textbf{Increment} &  \textbf{$\sigma$} \\
		\midrule
\multicolumn{1}{c|}{\multirow{2}{*}{$GD$}} & Closed & 14 & 18 &  -- & -- \tabularnewline 
\multicolumn{1}{c|}{} & Open  & 31 & 8 &  -- & -- \\  \hline 
\multicolumn{1}{c|}{\multirow{2}{*}{$QD^{25}_e$}} & Closed & 38 & 8 & $781\%$ & 1016\tabularnewline 
\multicolumn{1}{c|}{} & Open  & 45 & 11 & $46\%$ & 3\\  \hline 
\multicolumn{1}{c|}{\multirow{2}{*}{$QD^{50}_e$}} & Closed & 48 & 8 & $26\%$ & 8\tabularnewline 
\multicolumn{1}{c|}{} & Open  & 53 & 17 & $17\%$ & 8\\  \hline 
\multicolumn{1}{c|}{\multirow{2}{*}{$QD^{75}_e$}} & Closed & 54 & 8 & $14\%$ & 1\tabularnewline 
\multicolumn{1}{c|}{} & Open  & 56 & 16 & $6\%$ & 5\\ 
		\bottomrule
	\end{tabular}
	\end{scriptsize}
	\caption{
Average AP results in  $e_1$ and $e_2$  tested in different light conditions with respect to those used to qualify the detector (day/night time).}
	\label{tab:results_different_conditions}
\end{table}

\begin{table}[!htbp]
\vspace{-0.2cm}
\centering
\begin{scriptsize}
\begin{tabular}{c|ccccc}
\toprule
\textbf{Env.} & \textbf{Exp.} & $GT$ & $TP$ ($TP_{\%}$) & $FP$  ($FP_{\%}$) & $BFD$ ($BFD_{\%}$)  \\
\midrule
\multirow{4}{*}{$e_1$} &$GD$ & 1079 & 334 (30\%) & 56 (5\%) & 150 (13\%) \\ 
 & $QD^{25}_e$ & 1079 & 532 (49\%) & 62 (5\%) & 306 (28\%) \\ 
 & $QD^{50}_e$ & 1079 & 572 (53\%) & 65 (6\%) & 299 (27\%) \\ 
 & $QD^{75}_e$ & 1079 & 634 (58\%) & 56 (5\%) & 248 (22\%) \\ 
[2pt]\hline 
\multirow{4}{*}{$e_2$} &$GD$ & 1051 & 335 (31\%) & 68 (6\%) & 276 (26\%) \\ 
 & $QD^{25}_e$ & 1051 & 584 (55\%) & 55 (5\%) & 357 (33\%) \\ 
 & $QD^{50}_e$ & 1051 & 690 (65\%) & 40 (3\%) & 217 (20\%) \\ 
 & $QD^{75}_e$ & 1051 & 700 (66\%) & 48 (4\%) & 236 (22\%) \\
 \bottomrule
\end{tabular}
\end{scriptsize}
\caption{Extended results in $e_1$ and $e_2$ tested in different light conditions with respect to those used to qualify the detector (day/night time).}
\label{tab:metric_complete_different_light}
\end{table}

\begin{figure}[!htbp]
	\centering
    \includegraphics[width=0.32\linewidth]{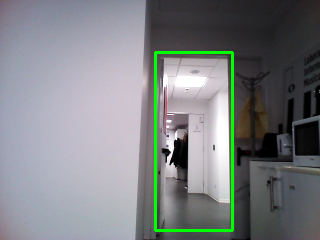}
	\hfill
    \includegraphics[width=0.32\linewidth]{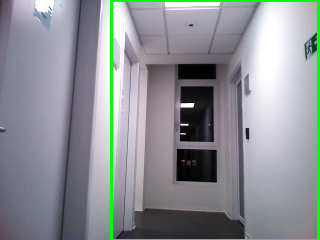}
    \hfill
    \includegraphics[width=0.32\linewidth]{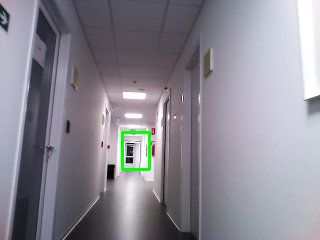}
    \\\vspace{0.15cm}
	\includegraphics[width=0.32\linewidth]{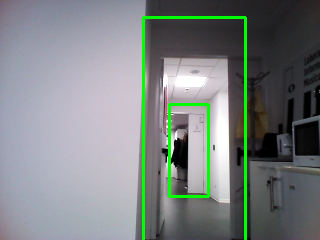}
	\hfill
	\includegraphics[width=0.32\linewidth]{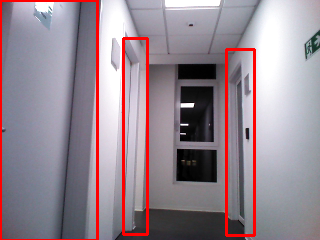}
	\hfill
	\includegraphics[width=0.32\linewidth]{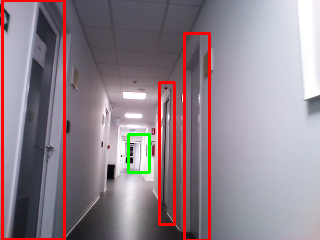}
	\caption{Challenging nighttime examples classified by $GD$ (top row) and $QD^{25}_e$ (bottom row). $QD^{25}_e$ is fine--tuned with examples obtained with daylight.}
	\label{fig:examples_different_conditions}
\end{figure}

The average AP obtained with different lighting conditions is reported in Table~\ref{tab:results_different_conditions} while the results of our extended metric (presented in Section~\ref{sec:metrics_real}) are shown in Table~\ref{tab:metric_complete_different_light}. Comparing them with Tables~\ref{tab:all_results_real_world}~and~\ref{tab:metric_complete} respectively, we can see how the performances of the $GD$ are robust to illumination changes, as they are similar to those obtained during daytime. More interestingly, it can be seen how the improvement of $QDs$ from the fine--tune is maintained also with different light conditions, with a slight performance decrease if compared to the results of Tables~\ref{tab:all_results_real_world}~and~\ref{tab:metric_complete}. This is a direct consequence of the fine--tune, which produces $QDs$ that slightly overfit the illumination conditions seen during training. Despite this, our method ensures a performance improvement to the $GD$ when used in long--term scenarios with illumination changes, enabling the $QDs$ to still solve challenging examples, as shown in Fig.~\ref{fig:examples_different_conditions}. Once again, $QD^{25}_{e}$, albeit using a few examples for fine--tuning, ensures the best performance improvement also under variable light conditions. See the video attachment, also linked in the repository, for additional examples of our method.

\section{Conclusions}

In this work, we presented a door--status detection method for mobile robots. Our method, based on a deep learning architecture, allows robots to recognise open or closed doors in challenging situations. To train our model, we built a dataset of labelled images from photorealistic simulations taking into account the point of view of a mobile robot.
We then fine--tuned a general model into a qualified one to increase performance in the robot's working environment. 

Future work will investigate how to quantify and reduce the effort needed for labelling examples to qualify a general detector. %
Furthermore, we will investigate online fine--tuning methods towards the goal to have a robot that can learn with experience to better distinguish features in its environment.%

\addtolength{\textheight}{-8.75cm}

\bibliographystyle{IEEEtran}
\bibliography{./contents/citations}
\end{document}